\documentclass[runningheads]{llncs}
\usepackage[T1]{fontenc}
\usepackage{amsmath} 
\usepackage{booktabs}
 \usepackage{multirow}
 \usepackage{xcolor}

\usepackage{hyperref}
\usepackage{graphicx}
\usepackage{url}

\begin{document}
\title{Minimizing the Effect of Sleep Deprivation in the Forward-Forward Algorithm}
\titlerunning{Minimizing the Effect of Sleep Deprivation in the FF Algorithm}
%
\author{Joy Datta\inst{1} \and
Puja Saha\inst{2} \and Rawhatur Rabbi\inst{1} \and Nafiz Imtiaz Rafin\inst{1} \and Swakkhar Shatabda\inst{1} \and Md. Golam Rabiul Alam\inst{1} \and Chad Mourning\inst{3}}
\authorrunning{J. Datta et al.}
%

\institute{Brac University, Dhaka 1212, Bangladesh \\
\email{joy.datta@g.bracu.ac.bd, rawhatur.rabbi@gmail.com, imtiazrafin1@gmail.com}\\ \and Rajshahi University of Engineering and Technology, Rajshahi 6204, Bangladesh \and Ohio University, Athens, OH 45701, United States \\ \email{mourning@ohio.edu}}
 
%
\maketitle              
\let\oldthefootnote\thefootnote
\let\thefootnote\relax

\footnotetext{This version of the contribution has been accepted for
publication, after peer review, but is not the Version of Record and does not
reflect post-acceptance improvements or any corrections. The Version of Record
is available online at
\url{https://doi.org/10.1007/978-3-032-31933-3_40}.}

\let\thefootnote\oldthefootnote

\begin{abstract}
This paper addresses the challenge posed by sleep deprivation in the Forward-Forward algorithm, where separating the two passes in this algorithm and imbalancing the data processing in the passes is considered an imitation of the cognitive processes observed in humans suffering from sleep deprivation. Previous research has demonstrated that sleep deprivation in the Forward-Forward algorithm has a catastrophic effect on learning efficacy. To mitigate this issue, we explore several approaches; these include alternative activation, optimized loss function, and threshold tuning. To simulate periodic rest, we reduce the number of positive passes in alternating epochs, creating short break phases. We additionally investigate the potential of caffeine-induced stimulation to enhance performance during sleep-deprived conditions. Experimental evaluations conducted on the MNIST and Fashion-MNIST datasets demonstrate that these modifications improve accuracy under the context of sleep deprivation. For example, a 2\%-62\% accuracy gain is observed in a severe sleep deprivation setting (16 positive or awake periods and 1 negative or sleep period). The approaches also enhance the resilience of the algorithm and its alignment with the adaptive mechanisms of human cognition.
\keywords{Forward-Forward Algorithm \and Sleep Deprivation \and Backpropagation \and Biologically Plausible Learning \and REM Sleep}
\end{abstract}

\section{Introduction}

Training deep neural networks typically requires minimizing a loss function by approximating gradients and updating parameters layer by layer using backpropagation~\cite{rumelhart1986learning}. While backpropagation has proven to be highly effective, it is considered biologically implausible due to its reliance on backward passes and non-local computations~\cite{lillicrap2020backpropagation}. Furthermore, the algorithm's high memory and computational overhead requirements pose practical limitations.

To address these concerns, the \textit{Forward-Forward (FF) algorithm} has emerged as a biologically inspired alternative that removes the backward pass entirely. Instead, FF uses two forward passes, one over positive (real) data and another over negative (synthetic or corrupted) data, to adjust weights based on a local goodness metric~\cite{hinton2022forward}. FF trains a model by keeping the sum of squared neural activities in layers above a threshold for positive data and well below that threshold for negative data. This approach aligns more closely with how learning is believed to occur in the brain, making it both computationally efficient and biologically plausible. FF also aligns with contrastive learning, a self-supervised learning framework where learning depends on distinguishing between similar and dissimilar data pairs. Although FF does not use an explicit contrastive loss, the dynamics of its dual-pass method share conceptual foundations with contrastive learning.

Interestingly, the structure of FF resonates with the mechanisms of human cognition, particularly the balance between \textit{waking experience} (the phase of new information processing) and \textit{REM sleep} (pruning irrelevant patterns). In the FF framework, the positive pass simulates learning from new experiences, while the negative pass plays the role of forgetting or pruning redundant patterns, similar to how REM sleep consolidates memory by suppressing irrelevant associations ~\cite{crick1983function,siegelrem}. When this balance is disrupted in humans, such as through sleep deprivation, cognitive decline is often observed~\cite{goel2009neurocognitive}.

This study explores solutions for sleep deprivation (SD) in FF, characterized by prolonged exposure to positive data without sufficient negative passes, which impairs the learning capacity of FF-based models. Geoffrey Hinton originally proposed the idea of separating the two passes to mimic natural sleep-wake cycles~\cite{hinton2022forward}. Building on this, a recent work formally studied the impact of such an imbalance, confirming that learning deteriorates when negative passes are underused ~\cite{licua2023sleep}. For example, accuracy drops from 74\% to only 10\% on MNIST when the negative pass is reduced from 16 to 1 while keeping the number of positive passes unchanged, just as humans underperform cognitively when sleep deprived~\cite{alhola2007sleep,lecun1998gradient}.

However, that work primarily observed the effect without offering solutions. In contrast, \textbf{our paper aims to both analyze and mitigate the impact of sleep deprivation in FF}. We approach this problem from three different perspectives:

    \textbf{Optimal Training Configurations}: We investigate the most effective activation function that outperforms previously used ReLU by a large margin and threshold configurations across layers under sleep deprivation. Furthermore, we propose some modifications to the loss functions to enhance accuracy and overall performance. 
    
    \textbf{Structured Rest Periods}: Inspired by circadian rhythms, we introduce short breaks between epochs to simulate intermittent sleep phases and balance the model's exposure to positive and negative data. This helps to achieve an approximate gain in accuracy in various training settings. 
    
    \textbf{Caffeine-Induced Loss}: The Caffeine-Induced Loss (CIL) integrates concepts from caffeine threshold modulation in neuronal firing with homeostasis and lateral inhibition mechanisms. This loss function enhances the learning stability during sleep-deprived training by dynamically adjusting loss contributions. This approach demonstrates a substantial accuracy gain, even under severe sleep deprivation (i.e., 16 positive or awake periods and 1 sleep or negative period). 


To establish both the biological and practical relevance of our findings, we conduct extensive experiments on MNIST and Fashion-MNIST, demonstrating that our proposed methods significantly improve the model accuracy in various sleep deprivation scenarios \cite{lecun1998gradient,xiao2017fashion}. These results highlight the potential of incorporating biologically inspired learning schedules and neural dynamics into AI systems, particularly in domains like continual or online learning where generating negative data on demand may be challenging.

\section{Related Works}

Deep learning has tremendous influence in the current world due to its applicability in a wide range of real-world scenarios, such as healthcare, e-commerce, and agriculture \cite{sarker2021machine}.  Current techniques usually rely on procedures that remain implausible to human cognitive processes, particularly in deep models that use the backpropagation algorithm, which propagates error derivatives backward to facilitate parameter update \cite{rumelhart1986learning}. While effective, it raises concerns over computation demand and memory requirements, as well as its misalignment with the mechanisms of cortical learning. Backpropagation has a lack of local plasticity, which means that parameter update requires information that is not locally available; additionally, backpropagation has no autonomy over the neural network. For example, prediction and learning stages require changes to the synaptic plasticity rules, whereas learning in the cortex works autonomously \cite{song2020can}.

The FF algorithm, a deep learning procedure, eliminates these issues by using two forward passes, instead of one forward and one backward pass, during backpropagation. One forward pass works on positive or desired data and a second on negative or undesired data, focusing on distinguishing these two types of input. The goal during learning is to maximize a goodness function for positive data and minimize it for the negative data without the need to propagate error derivatives backward, offering a more biologically plausible alternative to traditional backpropagation. It also has the advantage that it still can be used even if the precise details of the forward computation is unknown \cite{hinton2022forward}.

Recent research in biologically inspired learning has also investigated alternatives such as feedback alignment, predictive coding, and local Hebbian-like rules, all of which aim to bridge the gap between artificial and cortical learning ~\cite{lillicrap2016random,whittington2019theories,richards2019deep}. These studies align with the Forward-Forward framework in seeking local, biologically plausible alternatives to error backpropagation. In particular, Hebbian learning offers a local, biologically grounded alternative, following the principle \textit{`neurons that fire together, wire together'}. Recent implementations in neural networks, particularly when combined with Winner-Take-All competition and lateral inhibition, have demonstrated competitive performance in image classification tasks, promoting sparse hierarchical representations ~\cite{nimmo2025advancing}. Reward-modulated Hebbian plasticity has also been shown to support context-dependent learning, consistent with cortical synaptic mechanisms~\cite{miconi2017biologically}.

Recent studies have explored biologically plausible alternatives to traditional backpropagation to make artificial learning systems more aligned with how the human brain learns. A paper introduced the Contrastive Hebbian Learning with Random Feedback Weights (CHL-RF) algorithm, which replaces symmetric weight transport with fixed random feedback connections, a concept drawn from feedback alignment theories \cite{luo2023contrastive}. This model maintains local Hebbian updates while achieving competitive performance with backpropagation, making it a strong candidate for biologically inspired learning. CHL-RF operates via contrastive phases, distinguishing between positive and negative patterns, and shows convergence under mild assumptions, thereby contributing theoretical and empirical support to the design of biologically grounded models.

Adequate sleep is crucial for cognitive functioning in humans \cite{ramar2021sleep}. Rapid eye movement (REM) sleep has a crucial role in brain development, sensorimotor function, and memory consolidation across a diverse range of species. REM sleep helps to suppress redundant or unwanted patterns that consolidate learning \cite{crick1983function}. Dysfunction in REM sleep can lead to several disorders, including narcolepsy, which makes people very drowsy during the day \cite{peever2017biology}. 

The possibility of separation of two passes in the FF algorithm can potentially make the algorithm more biologically feasible to be implemented in the brain \cite{hinton2022forward}. Eliminating the negative phase updates for a while would mimic the catastrophic effect of severe sleep deprivation. Here, the positive pass resembles the waking phase, and the negative pass resembles the sleep phase in humans. During the positive pass, the neural network sees new information, and during the negative pass, it learns to avoid unwanted patterns (similar to human cognition). The separation of two passes and the imbalance of positive and negative phase updates affect learning, mimicking sleep deprivation in humans. Prolonged exposure to positive data compared to a negligible amount of exposure to negative data severely affects the accuracy of the model. Learning is also negatively affected when a large amount of positive data is followed by the same large amount of negative data. These findings suggest the importance of exploring the optimal phase balance and the possibility of using some other activation, the optimal threshold, or a better loss function in the FF algorithm \cite{licua2023sleep}.

Complementing this direction, researchers extended the FF algorithm beyond image recognition to natural language processing tasks like IMDb sentiment classification \cite{gandhi2023extending}. They proposed a pyramidal thresholding technique to improve FF's performance and stability. Their results underscore the importance of phase scheduling and hyperparameter optimization when adapting FF to new domains. These findings are particularly relevant for understanding how learning may proceed in systems where error backpropagation is not feasible.


While these studies emphasize biologically inspired alternatives, including FF that leverage local plasticity, contrastive learning, and non-backpropagation-based parameter updates, the effect of imbalanced learning phases is not well understood within the FF algorithm. This motivates our experiments with cognitively grounded principles to improve the robustness and accuracy of FF-based deep neural networks under sleep deprivation.

\section{Methodology}

\begin{figure}[htbp]
\vskip 0.1in
\begin{center}
\centerline{\includegraphics[width=0.61\columnwidth]{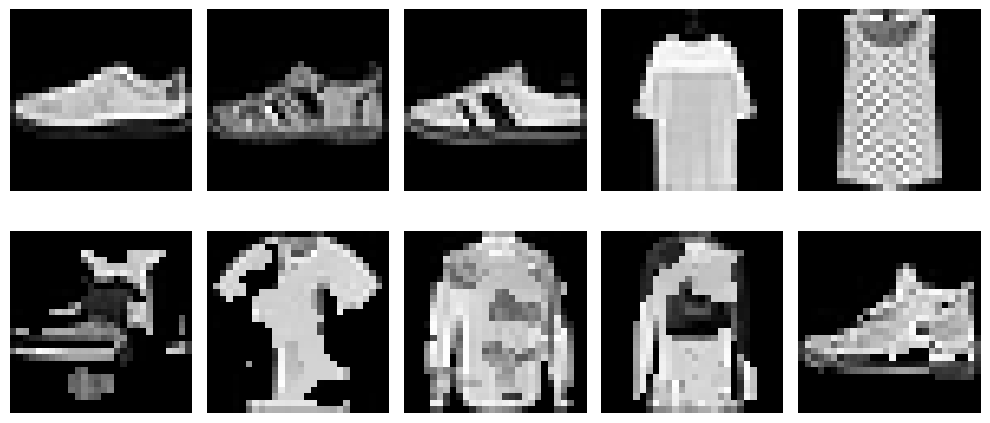}}
\caption{Positive (upper) and negative (lower) samples from Fashion MNIST.}
\label{fig:fmnist_samples}
\end{center}
\vskip -0.1in
\end{figure}

In this study, MNIST and Fashion-MNIST are used to assess the impact of sleep deprivation and to test possible solutions to this phenomenon in the FF algorithm \cite{lecun1998gradient,xiao2017fashion}. As the previous study on sleep deprivation in FF suggests that masking helps in sleep-deprived settings in the algorithm, we incorporate the same masking technique to generate negative data \cite{licua2023sleep}. The masking strategy follows a mixture of two data points combined using a predefined mask. Figure \ref{fig:fmnist_samples} shows the positive and negative data (generated using the masking strategy). Eq. \ref{eq:base_formula} shows the unchanged loss functions for positive pass and negative pass, respectively. Here, $\mathcal{L}_{pos}$ (loss for positive data) is responsible for maximizing layer goodness, whereas $\mathcal{L}_{neg}$ (loss for negative data) works to minimize it. It penalizes when $\mathcal{G}_{pos}<\mathcal{T}$ or $\mathcal{G}_{neg}>\mathcal{T}$.



\begin{equation}
\mathcal{L}_{pos} = \text{Softplus}(-\mathcal{G}_{pos} + \mathcal{T}), \quad
\mathcal{L}_{neg} = \text{Softplus}(\mathcal{G}_{neg} - \mathcal{T})
\label{eq:base_formula}
\end{equation}

where $\text{Softplus}(x) = \ln(1 + e^x)$, $\mathcal{G}_{pos}$, $\mathcal{G}_{neg}$, and $\mathcal{T}$ denotes goodness over a positive batch, goodness over a negative batch, and threshold of current layer, respectively.

The deep model has three layers with 500 neurons each, using a threshold of 1.5 for every layer (for baselines only), as previously used. The optimal layer-wise thresholds are found later after careful hyperparameter tuning. Various activation functions are assessed for the layers, including Rectified Linear Unit (ReLU), Exponential Linear Unit (ELU), Sigmoid Linear Unit (SiLU), Hyperbolic Tangent (Tanh), Hyperbolic Tangent Shrinkage (Tanhshrink), Softplus, and Mish. The optimal hyperparameters (layer-wise thresholds and activation) are used for the experiments. Adam optimizer with a learning rate of 0.001 for the negative pass, and 0.001 divided by the value of the awake period is used as the learning rate for the positive pass \cite{diederik2014adam}. We used a batch size of 512 for all experiments.

            \subsection{Simple Weighted Loss}
We first introduce a weight term for both positive and negative losses that mitigates the sleep deprivation effect on the FF. Here are the losses:

\begin{equation}
\mathcal{L}_{pos} = \alpha \cdot \text{Softplus}(-\mathcal{G}_{pos} + \mathcal{T}), \quad
\mathcal{L}_{neg} = \beta \cdot \text{Softplus}(\mathcal{G}_{neg} - \mathcal{T})
\label{eq:sw}
\end{equation}

where $\alpha$ and $\beta$ are simply the weights of the positive and negative loss terms. In this experiment, $\alpha$ = 1/awake period and $\beta$ = 1 to ensure that negative passes have balanced significance during training compared to positive passes.

\subsection{Caffeine-Induced Stimulation}
The caffeine-induced loss (CIL) function is detailed below:

\begin{align}
    \mathcal{L}_{\text{pos}}^{\text{c}} &= \lambda \cdot \text{Softplus}\left(-\mathcal{G}_{\text{pos}} + \mathcal{T} - \frac{\mathcal{G}_{\text{neg}}}{\mathcal{G}_{\text{pos}}+\varepsilon}-m_{c}\right) \label{eq:caff_pos} \\
    \mathcal{L}_{\text{pos}}^{\text{h}} &= \eta \cdot (\mathcal{A} - \text{r})^2 \label{eq:homeo_pos} \\
    \mathcal{L}_{\text{pos}} &= \mathcal{L}_{\text{pos}}^{\text{c}} + \mathcal{L}_{\text{pos}}^{\text{h}} \label{eq:total_pos} \\
    \mathcal{L}_{\text{neg}}^{\text{c}} &= \psi \cdot \text{Softplus}\left(\mathcal{G}_{\text{neg}} - \mathcal{T} + \frac{\mathcal{G}_{\text{neg}}}{\mathcal{G}_{\text{pos}}+\varepsilon} + m_{c}\right) \label{eq:caff_neg} \\
    \mathcal{L}_{\text{neg}}^{\text{i}} &= \delta \cdot \text{ReLU}\left(\mathcal{G}_{\text{neg}} - \mathcal{G}_{\text{pos}} + m_{g}\right) \label{eq:inhibit_neg} \\
    \mathcal{L}_{\text{neg}} &= \mathcal{L}_{\text{neg}}^{\text{c}} + \mathcal{L}_{\text{neg}}^{\text{i}} \label{eq:total_neg}
\end{align}

Caffeine tends to lower the threshold for neuronal firing in the brain. In Eq. \ref{eq:caff_pos}, positive loss is modulated through a dynamic term, $\mathcal{G}_{neg}$ / $\mathcal{G}_{pos} + \varepsilon$, to suppress the noise and minimize the threshold a little to simulate caffeine effect. Here, $m_{c}$ is a constant offset, $\epsilon$ is a small value to prevent division by zero, and $\lambda$ is the weight for the loss term $\mathcal{L}^{c}_{pos}$. Homeostasis is a biological process of maintaining internal stability despite changes using self-regulating processes. In Eq. \ref{eq:homeo_pos}, $\mathcal{L}^{h}_{pos}$ does the same during sleep deprivation in the FF analogously, where $\mathcal{A}$ is the current neuron activity. $\eta . (\mathcal{A} - r)^2$ regularizes neuron activity by forcing it to adhere to an ideal activation $r$. This homeostatic plasticity term (weighted by $\eta$) keeps the average firing rate or mean activity around a set point $r$. $\mathcal{L}^{c}_{pos}$ and $\mathcal{L}^{h}_{pos}$ together gives the total positive pass loss $\mathcal{L}_{pos}$. 

Eq. \ref{eq:caff_neg} shows the modified negative loss term with the caffeine effect. This ensures stronger punishment when goodness $\mathcal{G}_{neg}$ on negative data is too high, leading to increased contrast between positive and negative data. $\mathcal{L}^{c}_{pos}$ and $\mathcal{L}^{i}_{neg}$ together ensures separability in the representation space, whereas $\mathcal{L}^{h}_{pos}$ and $\mathcal{L}^{i}_{neg}$ builds robustness under sleep deprivation. Neuronal inhibition is the process by which neuron activities are suppressed to balance excitation and maintain stability. In Eq. \ref{eq:inhibit_neg}, we try to implement that mechanism in the FF. Here, $\mathcal{L}^{i}_{neg}$ tries to enforce that $\mathcal{G}_{pos}$ is greater than $\mathcal{G}_{neg}$ by a margin $m_{g}$. The FF-based neural network suppresses its activity when negative data produces stronger activity than its positive counterpart. 

Collectively, the entire loss term mimics the caffeine effect on the FF-based model when sleep deprived. First, the losses: $\mathcal{L}^{c}_{pos}$ and $\mathcal{L}^{c}_{neg}$ are the modified loss functions with caffeine effect by introducing higher excitability. Second, homeostatic plasticity loss: $\mathcal{L}^{h}_{pos}$ prevents uncontrolled activity growth when the threshold is lowered due to caffeine effect. Third, the loss inspired by the inhibition mechanism: $\mathcal{L}^{i}_{neg}$ forces the model to make sure that the positive data produce much stronger activities than negative data. This helps in preserving a discriminative structure when excitability is increased due to caffeine. Together, these result in a biologically inspired loss function for FF-based training of deep models during sleep deprivation that gives competitive performance. 

The values of the hyperparameters in the CIL are kind of arbitrary (not perfect, but logical choices). In Eq. \ref{eq:caff_pos}, Eq. \ref{eq:homeo_pos}, and Eq. \ref{eq:total_pos}, the hyperparameter values are: $\lambda$ = 1.2, $\epsilon$ = $10^{-5}$, $\eta$ = 0.25 (mild and sufficient homeostatic stabilization, allowing meaningful learning), and $r$ = 0.15 (smaller $r$ causes under-activation while larger $r$ causes dense activations, leading to lower discriminability). Similarly, hyperparameters in Eq. \ref{eq:caff_neg}, Eq. \ref{eq:inhibit_neg}, and Eq. \ref{eq:total_neg} are: $\psi$ = 2.4 (doubled $\lambda$, biasing learning towards suppressing spurious negative activations and making negative updates more informative during sleep deprivation), $\delta$ = 1.5, $m_{g}$ = 0.15 (a small tolerance margin to prevent premature inhibition due to noise), and $m_{c}$ = 0 (this constant offset is optional, setting it to a positive value can lead to a small static increased excitability by lowering threshold a little). Although these values are informed choices, not tuned enough, they still give a solution where we get better accuracy even when the model is sleep deprived. Carefully tuning these hyperparameters can lead to even better performance. 

\subsection{Incorporating Short Breaks}

We conducted two training approaches to explore possible solutions to the imbalance between the awake and sleep phases. In the first experiment, each epoch consistently follows a pattern where the model is trained for 8 or 16 batches of positive data and then only 1 batch of negative data, repeating this sequence for the duration of each epoch. The goal of this experiment is to simulate a continuous learning process without enough rest. 

In the second experiment, we vary the ratio of positive and negative data processing across epochs to simulate a more dynamic sleep-wake cycle. For instance, in the first epoch, the model is trained on 2 batches of positive data followed by 1 batch of negative data, and in the second epoch, we reduce the ratio between the sleep and awake phase to 1 batch of positive data followed by 1 batch of negative data, ensuring that the model receives more frequent rest periods. This approach is to simulate rest between learning, which could help improve the performance of the model under conditions that mimic sleep deprivation.

\section{Experimental Results}

This section provides a detailed discussion of the experimental results. The results indicate that sleep deprivation in the FF algorithm has a profoundly detrimental effect on its learning capacity. However, the choice of activation function, optimal threshold,  incorporating short breaks between alternating epochs, and bio-inspired loss functions help to mitigate this effect, according to our experimental results. The findings are discussed in this section.

\subsection{Finding the Optimal Layer-wise Threshold}

To evaluate the impact of various thresholds and determine the optimal value, we experiment with six different threshold combinations across our three-layer network. Each configuration is tested over 100 epochs with three independent trials. Here, we use the same hyperparameter settings used in the only previous study on sleep deprivation in the FF, except for the layer-wise thresholds \cite{licua2023sleep}. The performance is then measured using the mean $\pm$ standard deviation (\%) of the test accuracies. The pyramidal threshold setting with a threshold value of layer 1 = 1.0, layer 2 = 1.25, and layer 3 = 1.5 achieves the highest accuracy (37.47\% $\pm$ 1.05\%) among all combinations. This gradual increase in threshold values aligns with the requirement for greater activation thresholds in the higher layers of the cortex. Table \ref{tab:threshold-ablation} shows the threshold configurations and accuracies after 3 independent trials on the Fashion-MNIST dataset. 

\begin{table}
\centering
\caption{Layer-wise threshold combinations and their test accuracies (\%) after training the model three times for 100 epochs.}
\label{tab:threshold-ablation}
\begin{tabular}{|l|l|l|l|}
\hline
Layer 1 & Layer 2 & Layer 3 & Accuracy (mean $\pm$ std) \\
\hline
1.0  & 1.0  & 1.0  & $36.88 \pm 1.23$ \\
1.5  & 1.5  & 1.5  & $33.17 \pm 1.10$ \\
2.5  & 2.5  & 2.5  & $26.88 \pm 4.44$ \\
1.0  & 1.25 & 1.5  & $37.47 \pm 1.05$ \\
1.0  & 1.5  & 2.0  & $37.10 \pm 0.39$ \\
1.0  & 2.0  & 3.0  & $36.11 \pm 1.06$ \\
\hline
\end{tabular}
\end{table}

\subsection{Exploration for Better Activation Functions}

\begin{figure}[htbp]
\vskip -0.2in
\begin{center}
\centerline{\includegraphics[width=0.70\columnwidth]{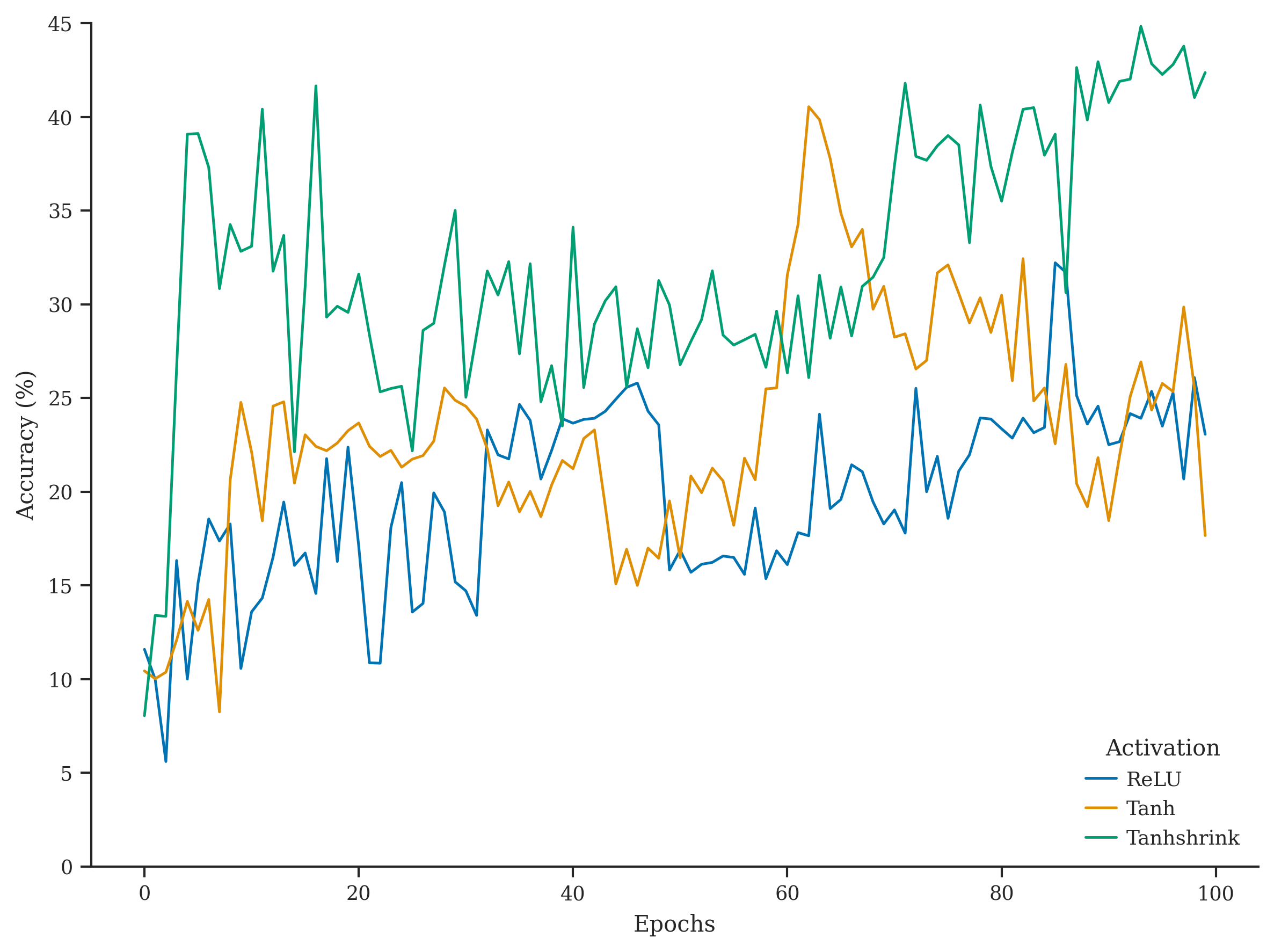}}
\caption{Performance of ReLU, Tanh, and Tanhshrink in the case of severe sleep deprivation (positive passes or awake period = 32, negative passes or sleep period = 1).}
\label{fig:pic1}
\end{center}
\vskip -0.5in
\end{figure}

In our second experiment, we try to find a better activation
 function that suits the FF algorithm in the case
 of sleep deprivation. Several activation functions are compared, including: ReLU, ELU, SiLU, Tanh, Tanhshrink, Softplus, Mish; we then find that when there is consistent sleep deprivation, Tanh and Tanhshrink works better than ReLU, which was used in the previous study \cite{licua2023sleep}. We train a model for 100 epochs (3 trials) on Fashion-MNIST, altering the activation function with the awake period equal to 8 and 16, and the sleep period equal to 1 at each epoch consistently, and find that ReLU, ELU, SiLU, and Softplus almost flatten out after a few epochs, whereas Tanh and Tanhshrink keep learning. But there are some strange movements in the Tanh and Tanhshrink learning curves, which seem to help to gain accuracy faster. Figure \ref{fig:pic1} compares the epochs versus accuracy curves for ReLU, Tanh, and Tanhshrink
 activation functions.


 Next, we find an alternative to ReLU
 that works better in the case of sleep deprivation, but when
 there are short breaks between alternating epochs. The model
 is again trained for 100 epochs on Fashion-MNIST, but at
 alternating epochs, the awake period toggles between 1 and 8 or 16 to ensure there are some short breaks. The activations, except Tanhshrink could not afford more than 37\% accuracy, whereas Tanhshrink achieves 51\%-53\%. Using the best activation function from the experiments, Tanhshrink, we compare the performance with the previous study in the subsequent subsections. Table \ref{tab:consistent_sleep} and \ref{tab:alternating_sleep} detail the study of finding the best activation for the consistent sleep deprivation and sleep deprivation with short breaks scenarios, respectively. 

\begin{table}[htbp]
\centering
\caption{Activation functions and their test accuracies (\%) after 100 epochs (consistent sleep deprivation setting).}
\label{tab:consistent_sleep}
\begin{tabular}{|l|l|l|}
\hline
Activation & Awake = 8 & Awake = 16 \\
\hline
ReLU        & $33.19 \pm 1.45$ & $25.23 \pm 3.06$ \\
ELU         & $41.81 \pm 1.28$ & $21.38 \pm 1.76$ \\
SiLU        & $38.00 \pm 0.58$ & $27.22 \pm 2.07$ \\
Tanh        & $49.15 \pm 1.09$ & $48.06 \pm 1.26$ \\
Tanhshrink  & $49.83 \pm 0.86$ & $48.63 \pm 1.48$ \\
Softplus    & $35.76 \pm 1.49$ & $27.81 \pm 2.90$ \\
Mish        & $42.59 \pm 0.80$ & $33.05 \pm 2.73$ \\
\hline
\end{tabular}
\end{table}


\begin{table}
\centering
\caption{Test accuracy (\%) on Fashion-MNIST under alternating awake--sleep pattern, mimicking short breaks after each epoch.}
\label{tab:alternating_sleep}
\begin{tabular}{|l|l|l|}
\hline
Activation & Awake = 8 & Awake = 16 \\
\hline
ReLU        & $25.54 \pm 0.26$ & $22.62 \pm 0.46$ \\
ELU         & $34.66 \pm 1.38$ & $23.26 \pm 0.51$ \\
SiLU        & $33.55 \pm 1.91$ & $34.93 \pm 1.96$ \\
Tanh        & $25.13 \pm 1.26$ & $16.36 \pm 0.48$ \\
Tanhshrink  & $53.60 \pm 1.00$ & $51.45 \pm 1.60$ \\
Softplus    & $24.60 \pm 1.34$ & $21.63 \pm 2.46$ \\
Mish        & $36.00 \pm 2.33$ & $37.15 \pm 1.49$ \\
\hline
\end{tabular}
\end{table}

\subsection{Modified Loss Functions}
\begin{figure}[!h]
  \centering
  \includegraphics[width=\textwidth]{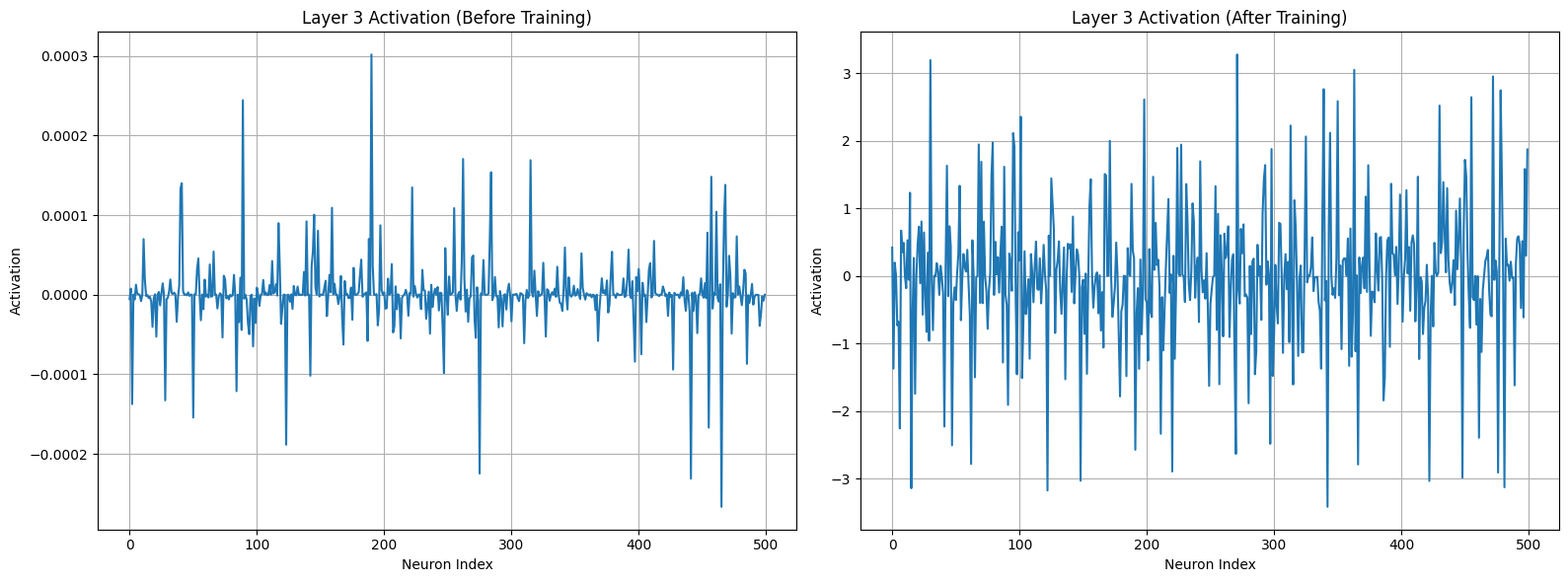}
  \caption{Layer activity when correct or positive data is fed. Random neural activity before training (left) and enhanced neural activity after training (right).}
  \label{fig:pic4}
\end{figure}

\begin{figure}[!h]
  \centering
  \includegraphics[width=\textwidth]{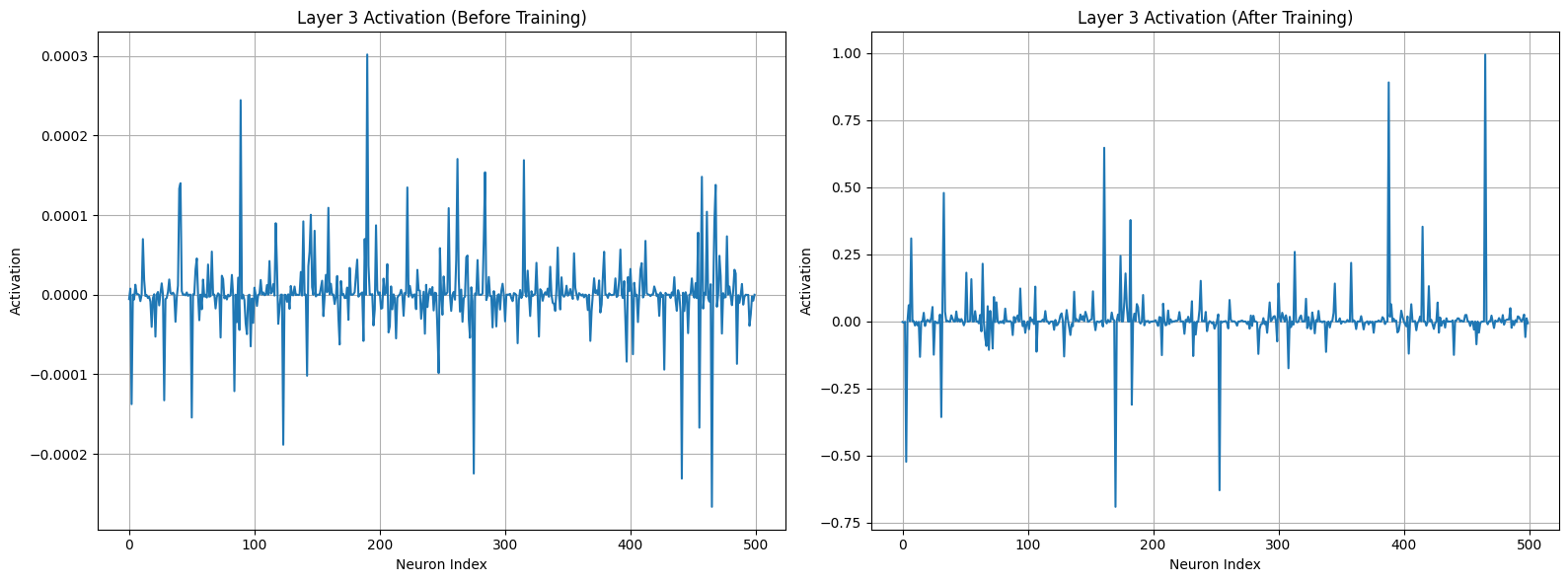}
  \caption{Layer activity when incorrect (e.g., image after applying the masking technique described) or negative data is fed. Random neural activity before training (left) and lowered neural activity. Since it has seen negative data, it must be below a certain threshold (right).}
  \label{fig:pic5}
\end{figure}

We first train the model on the datasets with the number of positive passes at each epoch ranging from 2 to 16, keeping the negative or sleep period equal to 1 to mimic sleep deprivation, like the previous study. To compare our modified loss functions with the baseline, we train the model twice for 500 epochs, first using the optimal configurations we found (Tanhshrink activation and optimal layer-wise threshold values) and the unchanged (simple weighted with $\alpha$ = 1, $\beta$ = 1) loss function, then using the caffeine-induced loss function. Table ~\ref{tab:table4} shows the results. It is evident from the table that during mild sleep deprivation, the unchanged or simple weighted loss function with the optimal threshold values and activation gives the best result. It is also noticeable that when awake or positive batches = 16 and sleep or negative batches = 1, CIL achieves 16\% increased accuracy than simple weighted loss combined with optimal configurations found. Figure \ref{fig:pic4} and Figure \ref{fig:pic5} show the layer activity before and after training. When the trained FF-based model witnesses positive data, it is excited to ensure that the sum of squared neural activities is above the predefined threshold, whereas witnessing negative data leads to inhibition in neurons.


\begin{table}
\centering
\setlength{\tabcolsep}{4pt}
\caption{Comparison of baseline and proposed methods on MNIST and Fashion-MNIST for different positive batch (awake) sizes.}
\begin{tabular}{|l|l|l|l|l|l|}
\hline
Dataset & Method & Awake=2 & Awake=4 & Awake=8 & Awake=16 \\
\hline
\multirow{3}{*}{MNIST}
 & Baseline & 78 & 75 & 73 & 10 \\
 & Optimal Configuration      & 75 & 77 & 75 & 56 \\
 & CIL      & 74 & 77 & 73 & 72 \\
\hline
\multirow{3}{*}{Fashion-MNIST}
 & Baseline & 55 & 54 & 52 & 49 \\
 & Optimal Configuration       & 52 & 54 & 75 & 68 \\
 & CIL      & 52 & 51 & 49 & 51 \\
\hline
\end{tabular}
\label{tab:table4}
\end{table}




\subsection{Intense and Prolonged Sleep Deprivation}
\begin{figure}[htbp]
  \centering
  \includegraphics[width=\textwidth]{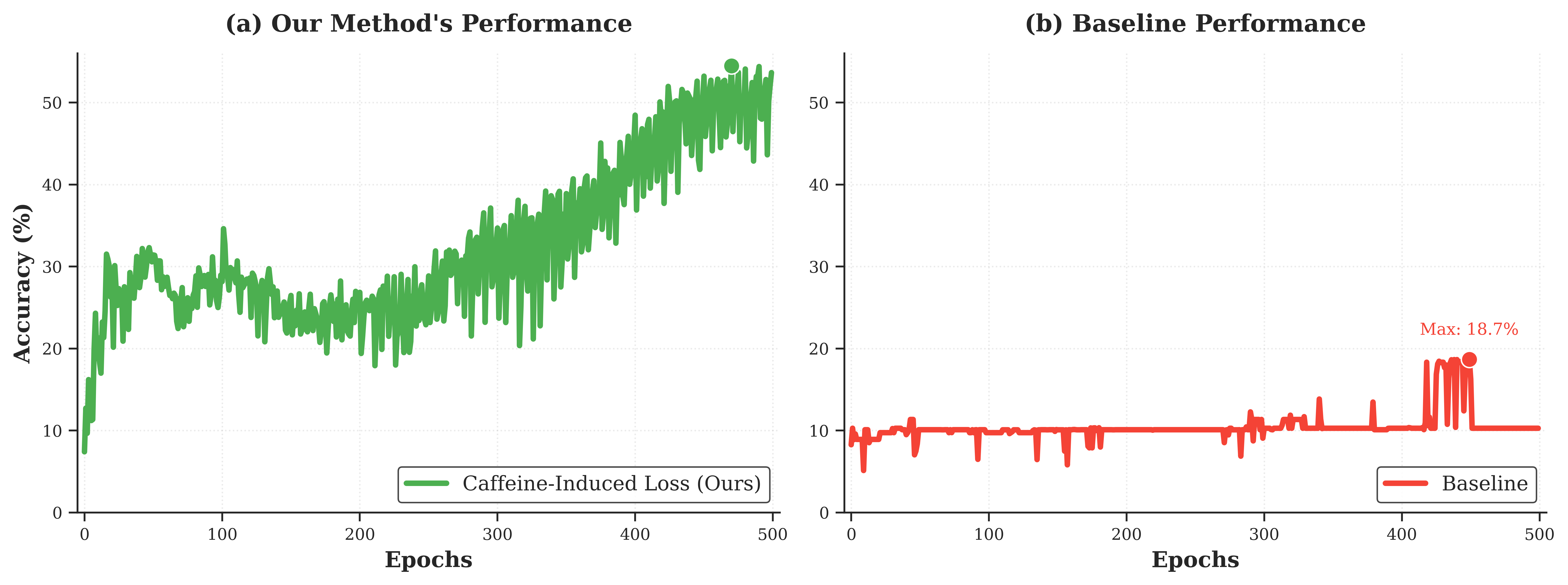}
  \caption{Epochs vs. accuracy curves of CIL and the baseline.}
  \label{fig:big1}
\end{figure}

Next, we experiment on severe sleep deprivation, meaning a large number of positive batches (awake period) ranging from 32 to 128, and only 1 batch of negative batches (sleep period) at each epoch during training. Here, we use the caffeine-induced loss function or CIL and train the model for 500 epochs. Since the previous study did not explore severe sleep deprivation settings (only explored up to 16 positive batches or awake periods), we train the model with the same configuration (for the baselines) used in that paper to validate our approach. Table \ref{tab:awake-positive-ablation} compares the baseline and our approaches. The weights for the simple weighted loss used are $\alpha$ = 1/awake period and $\beta$ = 1 are utilized, because we found that this particular choice of weights leads to better accuracy (results are discussed in the next subsection). Our caffeine-induced loss function and optimal training configurations combined give a significant accuracy boost in every severe sleep deprivation setting. In the MNIST experiment, we get at least a 20\% (positive batch = 128, negative batch = 1) to a maximum 36\% (positive batch = 64, negative batch = 1) gain in accuracy. Whereas in the Fashion-MNIST experiment, our approach enhances accuracy by 27\%\ to 28\%. Figure \ref{fig:big1} shows epochs vs. accuracy curve of our method and baseline on the MNIST dataset with positive batch (awake period) = 64 and negative batch (sleep period) = 1, over 500 epochs. Here, our approach consistently helps in learning, having an accuracy of around 55\%, whereas the baseline manages to achieve only around 19\% (accuracy oscillating between 10\% and 20\%).

\begin{table}
\centering
\setlength{\tabcolsep}{4pt}
\caption{Accuracy comparison across datasets, methods, and the number of awake or positive batches.}
\begin{tabular}{|l|l|l|l|l|}
\hline
Dataset & Method & Awake=32 & Awake=64 & Awake=128 \\
\hline
\multirow{3}{*}{MNIST}
 & Baseline & 29 & 19 & 21 \\
 & SW       & 30 & 20 & 17 \\
 & CIL      & 53 & 55 & 41 \\
\hline
\multirow{3}{*}{Fashion MNIST}
 & Baseline & 22 & 21 & 22 \\
 & SW       & 49 & 48 & 46 \\
 & CIL      & 50 & 49 & 49 \\
\hline
\end{tabular}
\label{tab:awake-positive-ablation}
\end{table}



\begin{figure}[!h]
\vskip 0.2in
\begin{center}
\centerline{\includegraphics[scale=0.28]{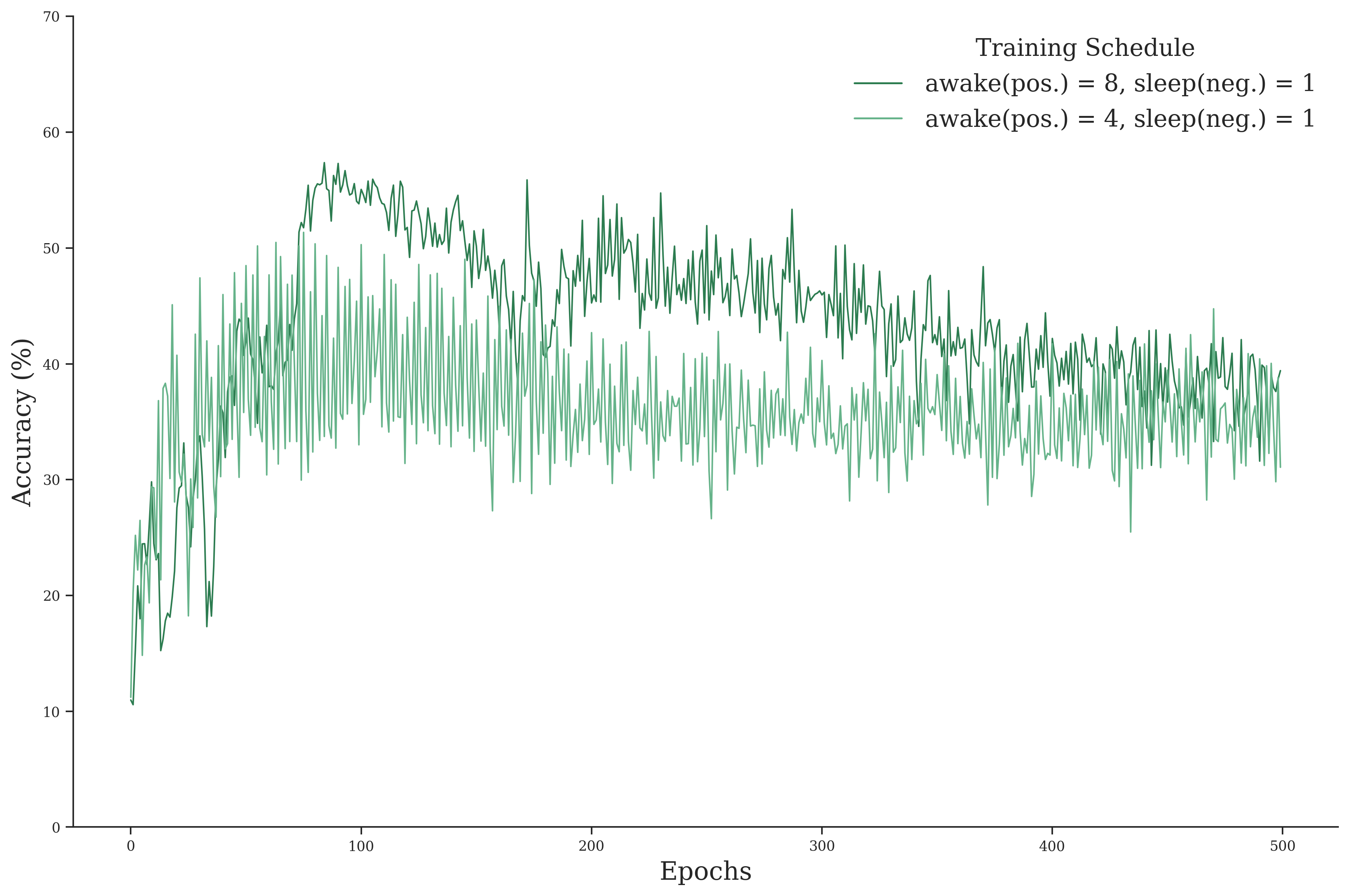}} 
\caption{Performance of ReLU, Tanh, and Tanhshrink in the case of severe sleep deprivation (awake period = 128, sleep period = 1).}
\label{fig:500_}
\end{center}
\vskip -0.2in
\end{figure}

It is evident from the epochs versus accuracy curve depicted in Figure \ref{fig:500_} that after approximately 100 epochs, accuracy starts to decrease, meaning that prolonged sleep deprivation might still affect learning despite a particular choice of activation. Although Tanhshrink works better in this regard, prolonged sleep deprivation still affects learning, but the choice of better activations helps to get better accuracy. Also, incorporating the caffeine-induced loss function helps to continue learning even after 100 or more epochs, as seen in Figure ~\ref{fig:big1}.


\subsection{Incorporating Short Breaks Between Alternating Epochs}
In the final experiment, we assess whether incorporating short
breaks can help mitigate the effects of sleep deprivation. We trained the model several times on Fashion-MNIST for 100
 epochs with an awake period ranging from 2 to 16, and a sleep period set at 1. But the awake period alternates between
 its value and 1 in alternating epochs to simulate short breaks. Table \ref{tab:alpha-beta-sd} compares performance during consistent sleep deprivation and sleep deprivation with short breaks with various $\alpha$ and $\beta$ values (weights for the positive and negative loss terms). It is clear from the table that with every $\alpha$ and $\beta$ choice, short breaks help to achieve better accuracy. Note that the training was performed for 100 epochs three times to validate that the results are statistically significant. Table \ref{tab:accuracy_comparison2} shows the comparison of accuracies when using consistent
 awake, sleep periods, and short breaks between the periods (this time, 500 epochs to let the model learn enough).
 It is evident from the results that incorporating short breaks
 helps to mitigate the effect of sleep deprivation in the FF algorithm, as the accuracies are much higher when
 there are short breaks. As learning goes, the model with consistent sleep and awake periods fluctuates around 50\% accuracy, whereas with short breaks, the
 model easily reaches more than 60\% accuracy without
 fluctuations.

\begin{table}
\caption{Comparison of test accuracy (\%) under consistent sleep deprivation (SD) vs.\ SD with short breaks using different values of $\alpha$ (awake period scale) and $\beta$ (sleep period scale).}
\centering
\begin{tabular}{|l|l|l|l|l|l|}
\hline
\multicolumn{2}{|c|}{} &
\multicolumn{2}{c|}{{Consistent SD}} &
\multicolumn{2}{c|}{{SD with Short Breaks}} \\
\hline
$\alpha$ & $\beta$ &
{Awake = 8} & {Awake = 16} &
{Awake = 8} & {Awake = 16} \\
\hline
1 & 1 & $49.83 \pm 0.86$ & $48.63 \pm 1.48$ & $53.06 \pm 1.00$ & $51.45 \pm 1.60$ \\
1 & 2 & $50.42 \pm 0.56$ & $50.68 \pm 0.63$ & $52.18 \pm 0.34$ & $50.17 \pm 0.44$ \\
1/awake period & 1 & $52.73 \pm 0.94$ & $49.21 \pm 0.61$ & $53.61 \pm 0.51$ & $52.48 \pm 1.63$ \\
\hline
\end{tabular}
\label{tab:alpha-beta-sd}
\end{table}

\begin{table}
\caption{Comparison of consistent awake and sleep periods vs.\ with short breaks.}
\label{tab:accuracy_comparison2}
\centering
\begin{tabular}{|l|l|l|}
\hline
No.\ of Pos.\ Batches & Consistent (\%) & Short Breaks (\%) \\
\hline
2  & 54.17 & 61.86 \\
4  & 45.18 & 49.52 \\
8  & 44.18 & 45.72 \\
16 & 38.63 & 43.31 \\
\hline
\end{tabular}
\end{table}

\section{Conclusion}
 In this paper, we have investigated the effects of sleep
 deprivation on the Forward-Forward algorithm and proposed
 strategies to mitigate its detrimental effects on the algorithm's
 learning efficacy. Our experimental results demonstrate that alternative activations, specifically Tanhshrink improves
 resilience to sleep deprivation effects compared to ReLU, that
 is used in the previous study. Moreover, incorporating structured breaks in the learning process also improves resilience
 of the algorithm, allowing the algorithm to better emulate
 the cognitive processes of humans. Introducing a caffeine-induced loss function further enhances the resilience and improved accuracy even under severe sleep deprivation settings in the algorithm. These findings contribute to
 the development of more adaptable and biologically plausible
 learning models in artificial intelligence.

%
%
%
\bibliographystyle{splncs04}
\bibliography{aaai2026}
%




\end{document}